\documentclass[10pt,twocolumn,letterpaper]{article}

\usepackage[pagenumbers]{cvpr}

\usepackage{graphicx}
\usepackage{amsmath}
\usepackage{amssymb}
\usepackage{booktabs}
\usepackage{multirow}

\usepackage[pagebackref,breaklinks,colorlinks]{hyperref}

\usepackage[capitalize]{cleveref}
\crefname{section}{Sec.}{Secs.}
\Crefname{section}{Section}{Sections}
\Crefname{table}{Table}{Tables}
\crefname{table}{Tab.}{Tabs.}

\begin{document}

\title{Equalized Focal Loss for Dense Long-Tailed Object Detection}

\author{Bo Li$^{1}$\footnotemark[1] \qquad Yongqiang Yao$^{2}$\footnotemark[1] \qquad Jingru Tan$^{1}$\footnotemark[1] \qquad Gang Zhang$^{3}$\\
Fengwei Yu$^{2}$ \qquad Jianwei Lu$^{1}$ \qquad Ye Luo$^{1}$\\
$^{1}$Tongji University \qquad $^{2}$SenseTime Research \qquad $^{3}$Tsinghua University\\
{\tt\small \{1911030,yeluo\}@tongji.edu.cn, \{soundbupt,tanjingru120\}@gmail.com}\\
{\tt\small yufengwei@sensetime.com, zhang-g19@mails.tsinghua.edu.cn, jwlu33@126.com}
}

\maketitle

\renewcommand{\thefootnote}{\fnsymbol{footnote}}
\footnotetext[1]{Equal contribution.}
\begin{abstract}
  Despite the recent success of long-tailed object detection, almost all long-tailed object detectors are developed based on the two-stage paradigm.
  In practice, one-stage detectors are more prevalent in the industry because they have a simple and fast pipeline that is easy to deploy. 
  However, in the long-tailed scenario, this line of work has not been explored so far.
  In this paper, we investigate whether one-stage detectors can perform well in this case. 
  We discover the primary obstacle that prevents one-stage detectors from achieving excellent performance is: categories suffer from different degrees of positive-negative imbalance problems under the long-tailed data distribution.
  The conventional focal loss balances the training process with the same modulating factor for all categories, thus failing to handle the long-tailed problem.
  To address this issue, we propose the Equalized Focal Loss (EFL) that rebalances the loss contribution of positive and negative samples of different categories independently according to their imbalance degrees. 
  Specifically, EFL adopts a category-relevant modulating factor which can be adjusted dynamically by the training status of different categories.
  Extensive experiments conducted on the challenging LVIS v1 benchmark demonstrate the effectiveness of our proposed method. With an end-to-end training pipeline, EFL achieves 29.2\% in terms of overall AP and obtains significant performance improvements on rare categories, surpassing all existing state-of-the-art methods.
  The code is available at \url{https://github.com/ModelTC/EOD}.
\end{abstract}

\section{Introduction}
\label{sec:intro}

\begin{figure}
  \centering
  \includegraphics[width=1.0\linewidth]{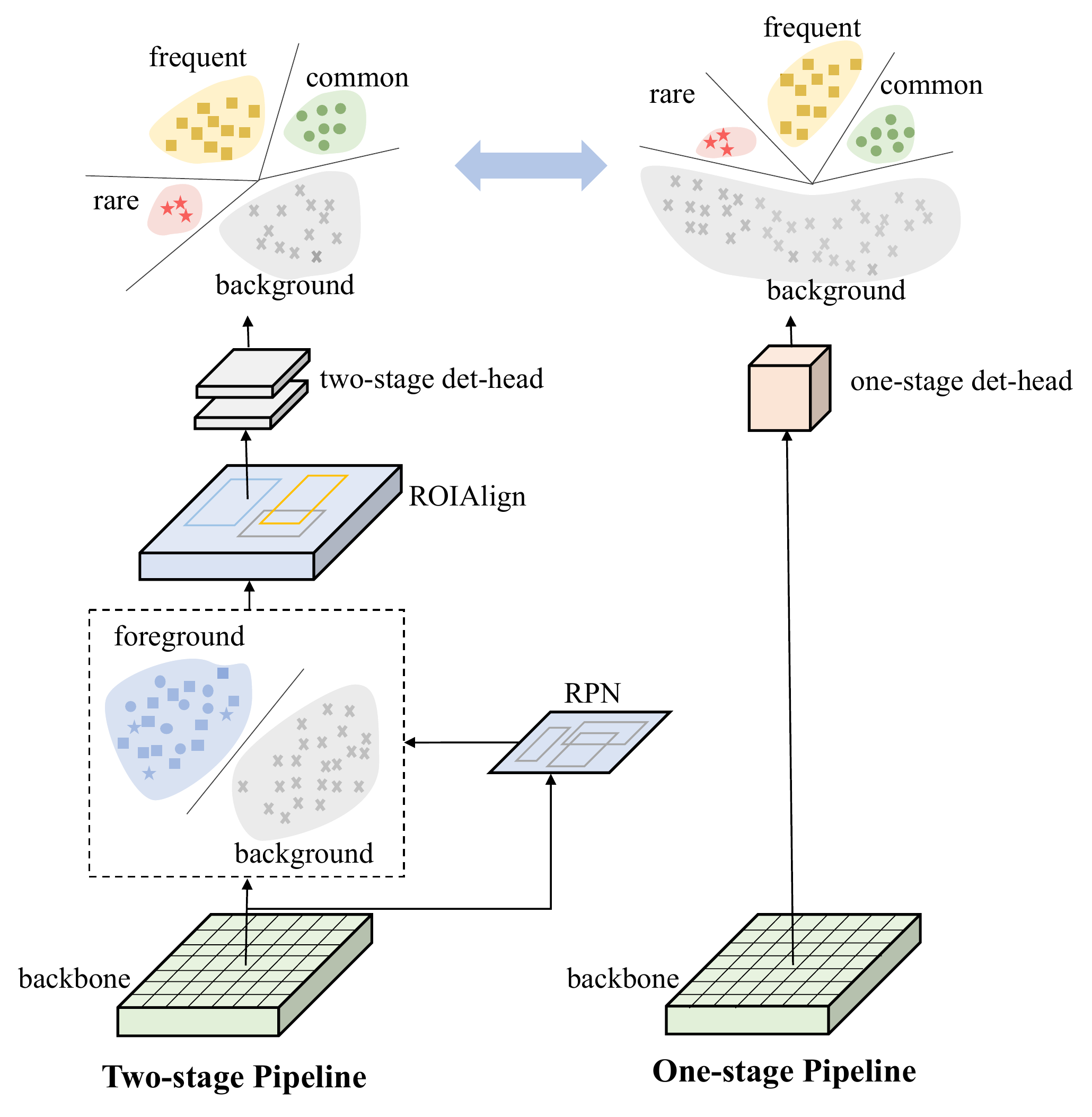}
  \caption{Visualization of the training pipeline in one-stage detectors and two-stage detectors.
  Different shapes indicate different categories, and their corresponding colors indicate the labels of the classifier at different stages.}
  \label{fig:pipeline}
\end{figure}

Long-tailed object detection is a challenging task that has received growing attention recently.
In the long-tailed scenario, data usually comes with a Zipfian distribution (\eg LVIS \cite{gupta2019lvis}) in which a few head classes contain plenty of instances and dominate the training process.
In contrast, a significant number of tail classes are instance-scarce thus perform poorly.
Common solutions to long-tailed object detection are data resampling \cite{gupta2019lvis, wang2020devil, wu2020forest, ren2020balanced, feng2021exploring}, decoupled training \cite{kang2019decoupling, wang2020devil, li2020overcoming}, and loss reweighting \cite{tan2020equalization, tan2021equalization, wang2021seesaw}.
Despite those successes in alleviating the long-tailed imbalance problem, almost all long-tailed object detectors are developed based on two-stage approaches popularized by R-CNN \cite{ren2015faster, he2017mask}.
In practice, one-stage detectors are more appropriate for a realistic scenario than two-stage ones because they are computationally efficient and easy to deploy.
However, there is no relevant work in this area.

In contrast to two-stage methods that incorporate a region proposal network (RPN \cite{ren2015faster}) to filter out most background samples before feeding proposals to the final classifier, one-stage detectors directly detect objects over a regular, dense set of candidate locations.
As presented in \cref{fig:pipeline}, the extreme foreground-background imbalance problem is introduced in one-stage detectors due to the dense prediction schema.
Combined with the foreground categories (\ie foreground samples of categories) imbalance problem in the long-tailed scenario, it severely damages the performance of one-stage detectors.

Focal loss \cite{lin2017focal} is a conventional solution to the foreground-background imbalance problem.
It focuses on the learning of hard foreground samples and reduces the impact of easy background samples with a specialized modulating factor. This loss redistribution technique works well under the category-balanced distribution but is inadequate to handle the imbalance problem among foreground categories in the long-tailed situation.
To solve this issue, we start from the existing solutions (\eg EQLv2 \cite{tan2021equalization}) in the two-stage pipeline and adapt them to work with focal loss together in one-stage detectors. Unfortunately, we find that these solutions only bring marginal improvements compared with their applications to two-stage detectors (see \cref{tab:eqlv2_baseline}).
Then we argue that simply combining existing solutions with focal loss can not address the two types of imbalance problems simultaneously.
By comparing the ratio of positive samples to negative samples in different data distributions (see \cref{fig:samples_number}), we further realize that the essence of these imbalance problems is that the positive-negative imbalance degrees are inconsistent among categories.
Rare categories suffer from more severe positive-negative imbalance than frequent ones, thus requiring more emphasis.

In this paper, we propose Equalized Focal Loss (EFL) by introducing a category-relevant modulating factor into focal loss.
The modulating factor with two decoupled dynamic factors (\ie the focusing factor and the weighting factor) deals with the positive-negative imbalance of different categories independently.
The focusing factor determines the concentration of learning on hard positive samples according to the imbalance degrees of their corresponding categories.
The weighting factor increases the impact of rare categories, ensuring that the loss contribution of rare samples will not be overwhelmed by frequent ones.
The synergy of these two factors enables EFL to overcome the foreground-background imbalance and the foreground categories imbalance uniformly when applying one-stage detectors to the long-tailed scenario.

We conduct extensive experiments on the challenging LVIS v1 \cite{gupta2019lvis} benchmark.
With a simple and effective one-stage training pipeline, EFL achieves 29.2\% AP, surpassing all existing long-tailed object detection methods.
Experimental results on OpenImages \cite{kuznetsova2020open} also demonstrate the generalization ability of our method.

To sum up, our key contributions can be summarized as follows:
(1) We are the first to study one-stage long-tailed object detection. We hope it will inspire the community to rethink the power and the value of one-stage detectors in the long-tailed scenario. 
(2) We propose a novel Equalized Focal Loss (EFL) that extends the original focal loss with a category-relevant modulating factor. It is a generalized form of focal loss that can address the foreground-background imbalance and foreground categories imbalance simultaneously.
(3) We conduct extensive experiments on LVIS v1 benchmark, and the results demonstrate the effectiveness of our approach. It establishes a new state-of-the-art and can be well applied to any one-stage detector.

\section{Related Work}
\label{sec:formatting}

\subsection{General Object Detection}

In recent years, benefiting from the great success of convolutional neural networks (CNN) \cite{krizhevsky2012imagenet, simonyan2014very, he2016deep, huang2017densely, radosavovic2020designing, hu2018squeeze}, the computer vision community has witnessed a remarkable improvement in object detection.
The modern object detection framework could be roughly divided into two-stage methods and one-stage methods.

\textbf{Two-stage methods.} With the emergence of Faster R-CNN \cite{ren2015faster}, two-stage methods \cite{girshick2014rich, girshick2015fast, ren2015faster, he2017mask, cai2018cascade, chen2019hybrid} occupy the dominant position in modern object detection.
Two-stage detectors first generate object proposals by a region proposal mechanism (\eg Selective Search \cite{uijlings2013selective}, or RPN \cite{ren2015faster}) and then perform spatial extraction of feature maps based on those proposals for further predictions.
Thanks to the proposal mechanism, plenty of background samples are filtered out.
Following \cite{ren2015faster}, the classifiers of most two-stage detectors are trained over a relatively balanced distribution of foreground and background samples, with a ratio of $1 \colon 3$.

\textbf{One-stage methods.} In general, one-stage detectors \cite{liu2016ssd, fu2017dssd, redmon2016you, redmon2017yolo9000, lin2017focal, law2018cornernet, duan2019centernet, tian2019fcos} have a simple and fast training pipeline which are closer to real-world applications.
In the one-stage scenario, detectors predict detection boxes from feature maps directly.
The classifiers of one-stage detectors are trained over a dense set with about $10^4$ to $10^5$ candidates, but only a few candidates are foreground samples.
Extensive studies \cite{liu2016ssd, shrivastava2016training, zhang2018single} attempt to address the extreme foreground-background imbalance problem from the hard example mining view or more complex resampling/reweighing schemes \cite{bulo2017loss}.
Focal loss \cite{lin2017focal} and its derivatives \cite{li2019gradient, li2020generalized,li2021generalized} reshape the standard cross-entropy loss such that they down-weight the loss assigned to well-classified samples and focus training on hard samples.
Benefit from the proposal of focal loss, one-stage detectors achieve very close performance to two-stage methods with a higher inference speed.
Recently, there are also some attempts \cite{zhang2020bridging, kim2020probabilistic, zhu2020autoassign, ge2021ota} to boost the performance from the label assignment perspective.
Our proposed EFL could be applied well with those one-stage frameworks and bring significant performance gains in the long-tailed scenario.

\subsection{Long-Tailed Object Detection}

Compared with general object detection, long-tailed object detection is a more complex task since it suffers from an extreme imbalance among foreground categories.
A straightforward solution to the imbalance is to perform data resampling during training.
Repeat factor sampling (RFS) \cite{gupta2019lvis} over-samples the training data from tail classes while under-samples those from head classes from the image level.
Wang \etal \cite{wang2020devil} train the detector in a decoupled way \cite{kang2019decoupling} and propose an extra classification branch with a class-balanced sampler from the instance level.
Forest R-CNN \cite{wu2020forest} resamples proposals from RPN with different NMS thresholds.
Other works \cite{ren2020balanced, feng2021exploring} implement data resampling through the meta-learning way or memory-augmented way.
Loss reweighting is another widely used solution to solve the long-tailed distribution problem.
Tan \etal \cite{tan2020equalization} propose the Equalization Loss (EQL) that alleviates the gradient suppression of head classes to tail classes.
EQLv2 \cite{tan2021equalization} is an upgraded version of EQL which adopts a novel gradient-guided mechanism to reweight the loss contribution of each category.
Wang \etal \cite{wang2021adaptive} interpret the long-tailed distribution problem from a statistic-free perspective and propose the adaptive class suppression loss (ACSL).
DisAlign \cite{zhang2021distribution} proposes a generalized reweighting method that introduces a balanced class prior to the loss design.
Besides data resampling and loss reweighting, many splendid works make attempts from different perspectives such as decoupled training \cite{wang2020devil, li2020overcoming}, margin modification \cite{ren2020balanced, wang2021seesaw}, incremental learning \cite{hu2020learning}, and causal inference \cite{tang2020long}.
Nevertheless, all those approaches are developed with two-stage object detectors, and there has been no relevant work on one-stage long-tailed object detection so far.
In this paper, we propose the first solution to the long-tailed object detection based on the one-stage pipeline.
It surpasses all existing methods in a simple and effective way.

\section{Methodology}
\label{sec:method}

\subsection{Revisiting Focal Loss}
\label{sec:revisiting_fl}

\begin{figure}
  \centering
  \includegraphics[width=1.0\linewidth]{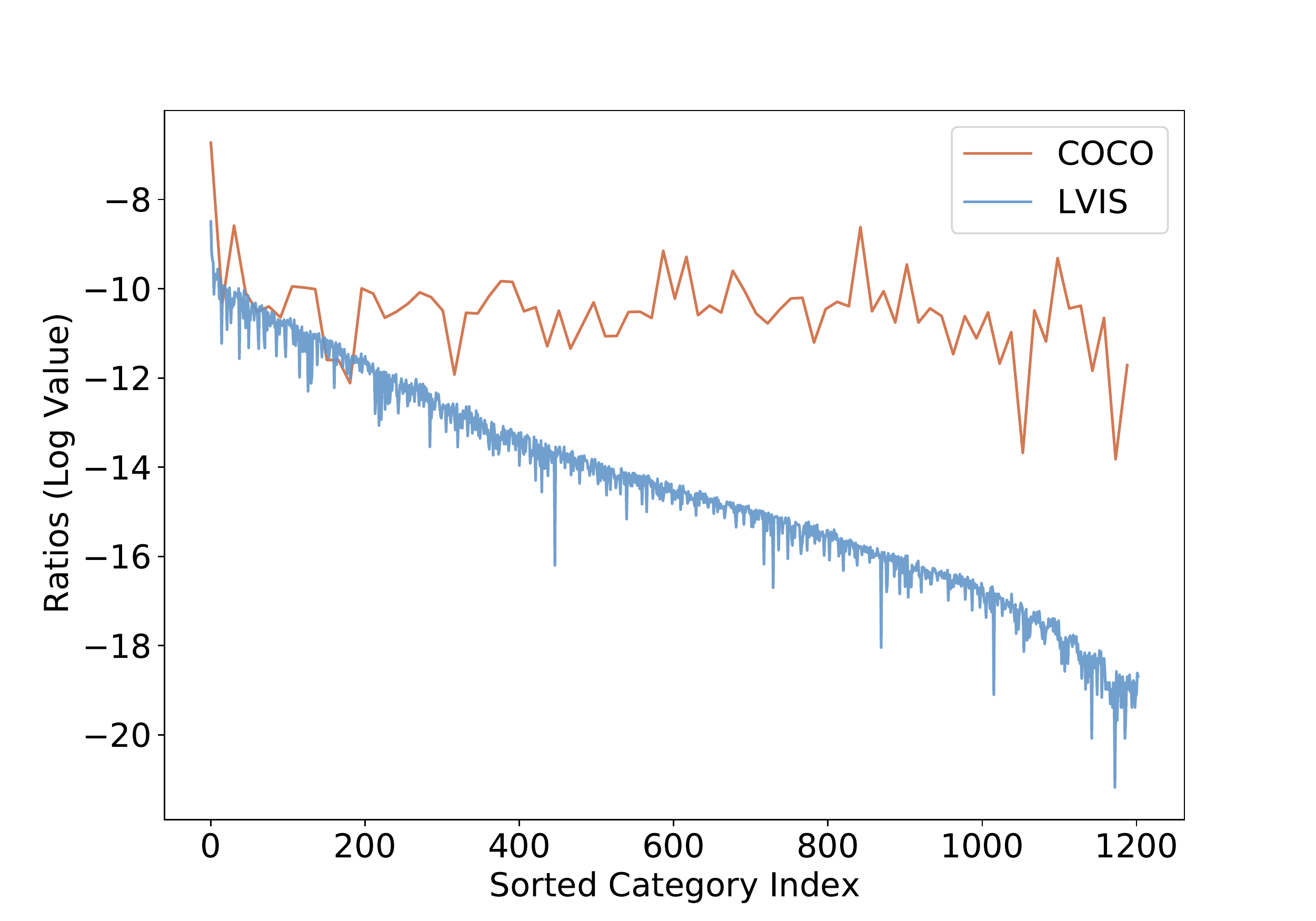}
  \caption{The ratio of the number of positive samples to negative samples on LVIS v1 \cite{gupta2019lvis} \texttt{train} split and COCO \cite{lin2014microsoft} \texttt{trainval35k} split.
  For better visualization, we demonstrate the log value of the ratio and align COCO 80 categories with LVIS 1203 categories.
  We adopt ATSS \cite{zhang2020bridging} as the sample selection strategy to distinguish between foreground samples and background samples.}
  \label{fig:samples_number}
\end{figure}

In one-stage detectors, focal loss \cite{lin2017focal}  is the widely-used solution to the foreground-background imbalance problem.
It redistributes the loss contribution of easy samples and hard samples, which greatly weaken the influence of the majority of background samples.
The formula of focal loss for binary classification is:
\begin{equation}
  \mathrm{FL}\left(p_{\mathrm{t}}\right)=-\alpha_{\mathrm{t}}\left(1-p_{\mathrm{t}}\right)^{\gamma} \log \left(p_{\mathrm{t}}\right)
  \label{eq:focal_loss}
\end{equation}

As declared in \cite{lin2017focal}, the term $p_{\mathrm{t}} \in \left[0, 1\right]$ indicates the predicted confidence score of an object candidate and the term $\alpha_{\mathrm{t}}$ is the parameter that balances the importance of positive samples and negative samples.
The modulating factor $\left(1-p_{\mathrm{t}}\right)^{\gamma}$ is the key component of focal loss.
It down-weights the loss of easy samples and focuses on the learning of hard samples through the predicted score $p_{\mathrm{t}}$ and the focusing parameter $\gamma$.
As mentioned in \cite{li2019gradient}, a large number of negative samples are easy to classify while positive samples are usually hard.
Hence, the imbalance between positive samples and negative samples could be roughly regarded as the imbalance between easy samples and hard samples.
The focusing parameter $\gamma$ determines the effect of focal loss.
It could be concluded from \cref{eq:focal_loss} that a big $\gamma$ will greatly reduce the loss contribution from most negative samples, thus improving the impact of positive samples.
This conclusion indicates that the higher the imbalance degree between positive samples and negative samples, the bigger the expected value of $\gamma$.

When it comes to the multi-class case, focal loss is applied to $C$ classifiers that act on the output logits transformed by the sigmoid function for each instance.
$C$ is the number of categories which means that a classifier is responsible for a specific category, \ie a binary classification task. 
Since focal loss equally treats the learning of all categories with the same modulating factor, it fails to handle the long-tailed imbalance issue (see \cref{tab:main_results}).

\subsection{Equalized Focal Loss Formulation}
\label{sec:efl_definition}

\begin{figure*}
  \centering
  \begin{subfigure}{0.49\linewidth}
    \includegraphics[width=1.0\linewidth]{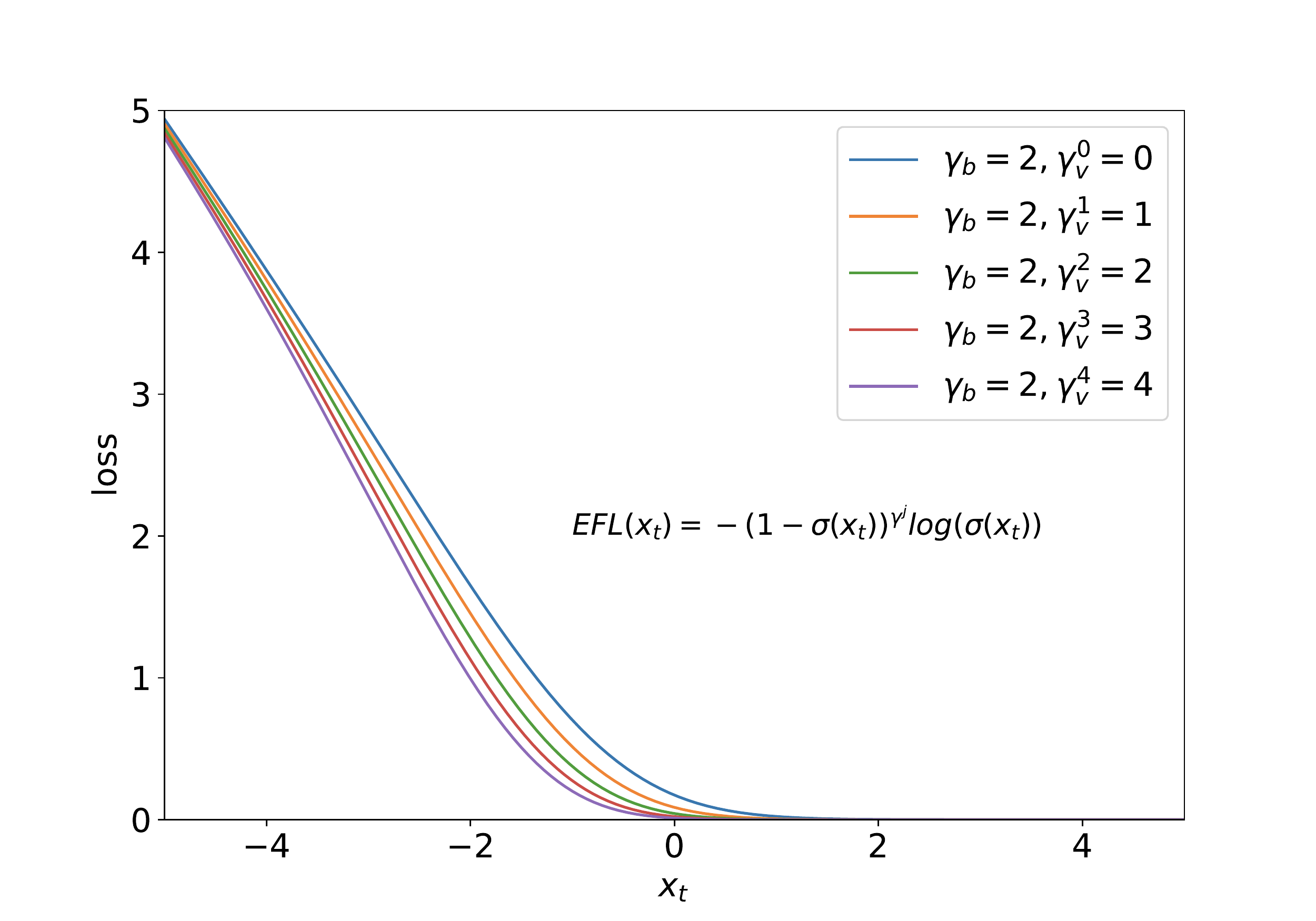}
    \caption{Equalized Focal Loss without the weight factor.}
    \label{fig:focal_loss}
  \end{subfigure}
  \hfill
  \begin{subfigure}{0.49\linewidth}
    \includegraphics[width=1.0\linewidth]{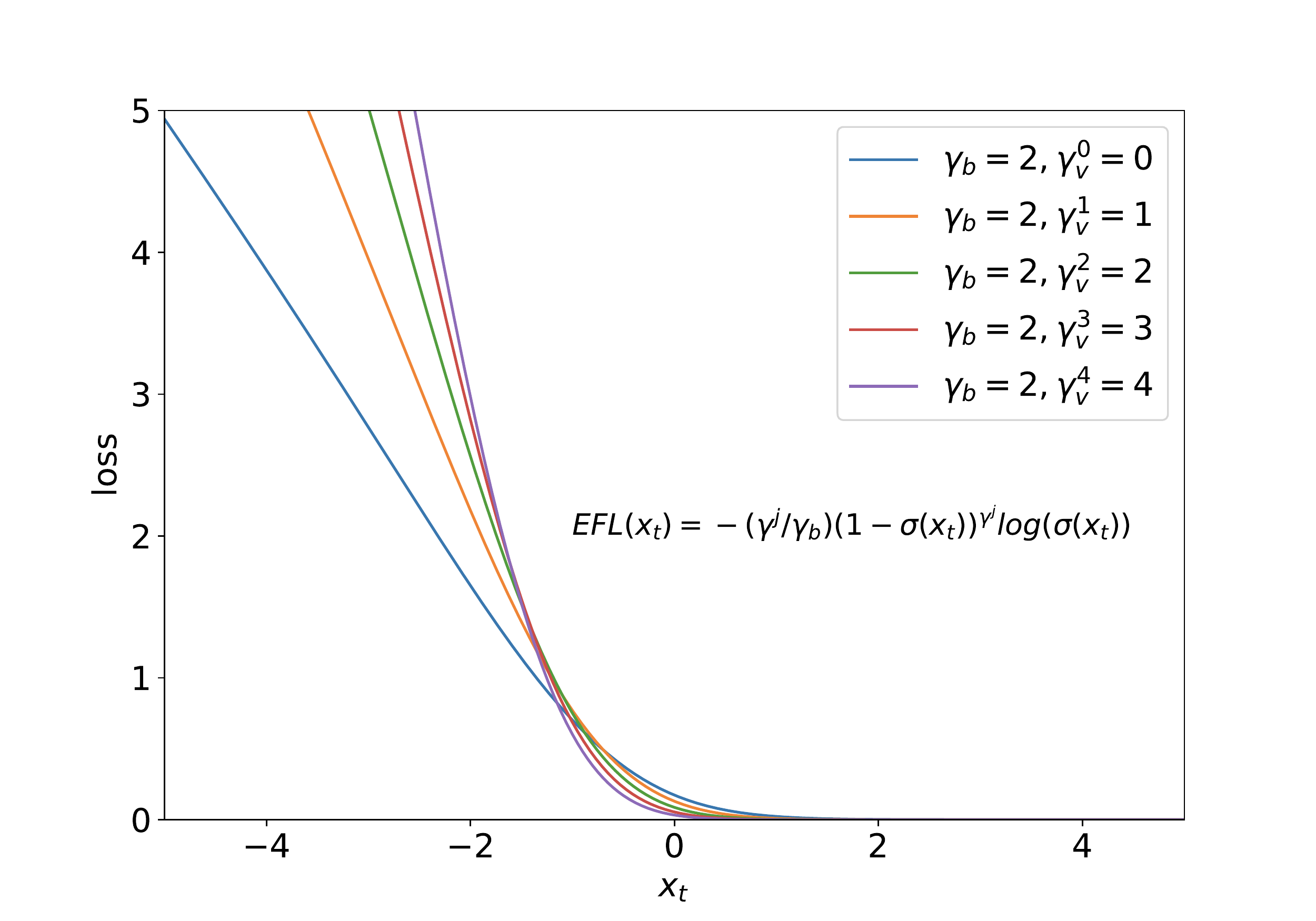}
    \caption{Equalized Focal Loss with the weight factor.}
    \label{fig:eq_focal_loss}
  \end{subfigure}
  \caption{Comparison of the loss contribution between EFL with (b) and without (a) the weighting factor.
  $x_{\mathrm{t}} = \left(2y - 1\right)x$, where $x$ is the output predicted logit and $y \in \{0, 1\}$ specific the ground-truth label of the binary classification.
  $\sigma$ indicates the sigmoid function.
  We set the $\gamma_b$ always to be equal to 2 and ignore the impact of $\alpha_{\mathrm{t}}$ in EFL.
  Different colors indicate different categories.}
  \label{fig:focal_loss_vis}
\end{figure*}

In the long-tailed dataset (\ie LVIS), in addition to the foreground-background imbalance, the classifier of a one-stage detector also suffers from the imbalance among foreground categories.
As shown in \cref{fig:samples_number}, if we view from the y-axis, the values of the ratio between positive samples to negative samples are much less than zero, which mainly reveal the imbalance between foreground and background samples. Here we refer to the value of the ratio as the positive-negative imbalance degree.
From the view of the x-axis, we can see that the imbalance degrees across categories show great differences, which indicates the imbalance among foreground categories.
Obviously, in the balanced data distribution (\ie COCO), the imbalance degree of all categories are similar. So it is enough to use the same modulating factor for all categories in focal loss. In contrast, those imbalance degrees are different in the long-tailed data situation. Rare categories suffer from more severe positive-negative imbalance than frequent ones. As presented in \cref{tab:eqlv2_baseline}, most one-stage detectors perform worse on rare categories than on frequent ones.
It indicates that the same modulating factor is not suitable for all positive-negative imbalance problems with different degrees.

\noindent
\textbf{Focusing Factor.} Based on the above analysis, we propose the Equalized Focal Loss (EFL) that adopts a category-relevant focusing factor to address the positive-negative imbalance of different categories separately.
We formulate the loss of the $j$-th category as:
\begin{equation}
  \mathrm{EFL}\left(p_{\mathrm{t}}\right)=-\alpha_{\mathrm{t}}\left(1-p_{\mathrm{t}}\right)^{\gamma^j} \log \left(p_{\mathrm{t}}\right)
  \label{eq:efl_loss}
\end{equation}
where $\alpha_{\mathrm{t}}$ and $p_{\mathrm{t}}$ are the same as that in focal loss.

The parameter $\gamma^j$ is the focusing factor for the $j$-th category, which plays a similar role as $\gamma$ in focal loss. As mentioned in \cref{sec:revisiting_fl}, different values of the $\gamma$ correspond to different degrees of positive-negative imbalance issues. We adopt a big $\gamma^j$ to alleviate the severe positive-negative imbalance problems in rare categories. And for frequent categories with slight imbalance, a small $\gamma^j$ is appropriate.
The focusing factor $\gamma^j$ is decoupled into two components, specifically a categories-agnostic parameter $\gamma_b$ and a categories-specific parameter $\gamma_v^j$:
\begin{equation}
  \begin{aligned}
  \gamma^j &=\gamma_b + \gamma_v^j \\
  &=\gamma_b + s\left(1 - g^j\right)
  \end{aligned}
  \label{eq:gamma_formula}
\end{equation}
where $\gamma_b$ indicates the focusing factor in a balanced data scenario that controls the basic behavior of the classifier.
And the parameter $\gamma_v^j \geq 0$ is a variable parameter associated with the imbalance degree of the $j$-th category.
It determines how much the concentration of learning is on positive-negative imbalance problems.
Inspired by EQLv2 \cite{tan2021equalization}, we adopt the gradient-guided mechanism to choose $\gamma_v^j$.
The parameter $g^j$ indicates the accumulated gradient ratio of positive samples to negative samples of the $j$-th category.
As mentioned in \cite{tan2021equalization}, a big value of $g^j$ indicates the $j$-th category (\eg frequent) is trained balance while a small value indicates the category (\eg rare) is trained imbalanced.
To satisfy our requirement of $\gamma^j$, we clamp the value of $g^j$ in the range $\left[0, 1\right]$ and adopt $1 - g^j$ to invert its distribution.
The hyper-parameter $s$ is a scaling factor that determines the upper limit of $\gamma^j$ in EFL.
Compared with focal loss, EFL handles the positive-negative imbalance problem of each category independently, which leads to the performance improvement (see \cref{tab:ablation_study_components}).

\noindent
\textbf{Weighting Factor.} Even with the focusing factor $\gamma^j$, there are still two obstacles damaging the performance:
(1) For a binary classification task, a bigger $\gamma$ is suitable to a more severe positive-negative imbalance problem. While in the multi-class case, as presented in \cref{fig:focal_loss}, for the same $x_{\mathrm{t}}$\footnote[2]{see the caption in \cref{fig:focal_loss_vis}.}, the bigger the value of $\gamma$, the smaller the loss. It leads to the fact that when we want to increase the concentration on learning a category with severe positive-negative imbalance, we have to sacrifice part of its loss contribution in the overall training process.
Such a dilemma prevents rare categories from achieving excellent performance.
(2) When $x_{\mathrm{t}}$ is small, the losses of samples from different categories with distinct focusing factors will converge to a similar value. Actually, we expect rare hard samples to make more loss contribution than frequent hard ones since they are scarce and couldn't dominate the training process.

We propose the weighting factor to alleviate the above problems by re-balancing the loss contribution of different categories. Similar to the focusing factor, we assign a big value of the weighting factor for rare categories to raise their loss contributions while keeping the weighting factor close to 1 for frequent categories.
Specifically, we set the weighting factor of the $j$-th category to $\frac{\gamma_b + \gamma_v^j}{\gamma_b}$ to be consistent with the focusing factor.
The final formula of EFL is:
\begin{equation}
  \mathrm{EFL}\left(p_{\mathrm{t}}\right)=-\sum_{j=1}^{C}\alpha_{\mathrm{t}}\left(\frac{\gamma_b + \gamma_v^j}{\gamma_b}\right)\left(1-p_{\mathrm{t}}\right)^{\gamma_b + \gamma_v^j} \log \left(p_{\mathrm{t}}\right)
  \label{eq:total_efl_loss}
\end{equation}
As shown in \cref{fig:eq_focal_loss}, with the weighting factor, EFL significantly increases the loss contribution of rare categories. Meanwhile, it focuses more on the learning of rare hard samples compared with frequent hard ones.

The focusing factor and the weighting factor make up the categories-relevant modulating factor in EFL. It enables the classifier to dynamically adjust the loss contribution of a sample according to its training status $p_{\mathrm{t}}$ and its corresponding category status $\gamma^j$.
As presented in \cref{sec:ablation_studies}, both focusing factor and weighting factor play significant roles in EFL.
Meanwhile, in the balanced data distribution, EFL with all $\gamma_v^j=0$ is equivalent to focal loss.
Such an appealing property makes EFL could be applied well with different data distributions and data samplers. 

\section{Experiments}

\subsection{Experimental Settings}

\noindent
\textbf{Dataset.} We perform our experiments on the challenging LVIS v1 \cite{gupta2019lvis} dataset.
LVIS is a large vocabulary benchmark for long-tailed object recognition, which contains 1203 categories.
Following the common practice \cite{tan2021equalization, wang2021seesaw}, all 100k images with about 1.3M instances in the \texttt{train} split are used for training, and all 20k images in the \verb+val+ split are used as validation for the analysis.
The categories in the LVIS v1 dataset are divided into three groups according to the numbers of images that appear in the \texttt{train} split: rare categories (1-10 images), common categories (11-100 images), and frequent categories ($>$100 images).

\begin{table}
  \centering
  \setlength{\tabcolsep}{2mm}{
  \begin{tabular}{l|cccc}
    \toprule
    method & AP & AP$_r$ & AP$_c$ & AP$_f$ \\
    \midrule
    RetinaNet \cite{lin2017focal} & 18.5 & 9.6 & 16.1 & 25.0 \\
    +EQLv2\&Focal & 20.5 (+2.0) & 12.6 & 19.2 & 25.4 \\
    \midrule
    FCOS$^{*}$ \cite{tian2019fcos} & 22.6 & 12.7 & 20.9 & 28.9 \\
    +EQLv2\&Focal & 23.0 (+0.4) & 14.1 & 21.3 & 28.7 \\
    \midrule
    PAA \cite{kim2020probabilistic} & 23.7 & 14.2 & 21.6 & 30.2 \\
    +EQLv2\&Focal & 24.1 (+0.4) & 16.5 & 22.1 & 29.8 \\
    \midrule
    ATSS \cite{zhang2020bridging} & 24.7 & 13.7 & 23.4 & 31.1 \\
    +EQLv2\&Focal & 25.2 (+0.5) & 15.0 & 24.3 & 30.8 \\
    \midrule
    Faster R-CNN \cite{ren2015faster} & 24.1 & 14.7 & 22.2 & 30.5 \\
    Seasaw Loss \cite{wang2021seesaw} & \textbf{26.4 (+2.3)} & \textbf{17.5} & \textbf{25.3} & \textbf{31.5} \\
    \bottomrule
  \end{tabular}}
  \caption{Results of different one-stage detectors with the combination of focal loss and EQLv2 \cite{tan2021equalization} on LVIS v1 benchmark.
  EQLv2\&Focal indicates this combination.
  All detectors are trained with ResNet-50-FPN by a 2x schedule using a repeat factor sampler.
  FCOS$^{*}$ indicates that the reported FCOS result is trained with a center-sampling strategy \cite{zhang2020bridging}.}
  \label{tab:eqlv2_baseline}
\end{table}

\noindent
\textbf{Evaluation Metric.} The results of object detection are evaluated with the widely-used metric AP of boxes prediction across IoU thresholds from 0.5 to 0.95.
Additionally, we report AP$_r$, AP$_c$, and AP$_f$, which indicate the boxes AP for rare, common, and frequent categories, respectively.

\begin{table*}
  \centering
  \setlength{\tabcolsep}{3.5mm}{
  \begin{tabular}{c|l|ccc|cccc}
    \toprule
    backbone & method & strategy & sampler & epoch & AP & AP$_r$ & AP$_c$ & AP$_f$ \\
    \midrule
    \multirow{11}{*}{ResNet-50} & \textit{two-stage} &  &  &  &  &  &  &  \\
    ~ & Faster R-CNN \cite{ren2015faster} & end-to-end & RFS & 24 & 24.1 & 14.7 & 22.2 & 30.5 \\
    ~ & EQL \cite{tan2020equalization} & end-to-end & RFS & 24 & 25.1 & 15.7 & 24.4 & 30.1 \\
    ~ & EQLv2 \cite{tan2021equalization} & end-to-end & RFS & 24 & 25.5 & 16.4 & 23.9 & 31.2 \\
    ~ & Seasaw Loss \cite{wang2021seesaw} & end-to-end & RFS & 24 & 26.4 & 17.5 & 25.3 & 31.5 \\
    ~ & cRT \cite{kang2019decoupling} & decoupled & RFS+CBS & 24+12 & 24.8 & 15.9 & 23.6 & 30.1 \\
    ~ & BAGS \cite{li2020overcoming} & decoupled & RFS+CBS & 24+12 & 26.0 & 17.2 & 24.9 & 31.1 \\
    \cmidrule{2-9}
    ~ & \textit{one-stage} &  &  &  &  &  &  &  \\
    ~ & RetinaNet \cite{lin2017focal} & end-to-end & RFS & 24 & 18.5 & 9.6 & 16.1 & 25.0 \\
    ~ & Baseline$^{\dag}$ & end-to-end & RFS & 24 & 25.7 & 14.3 & 23.8 & \textbf{32.7} \\
    ~ & EFL (Ours) & end-to-end & RFS & 24 & \textbf{27.5} & \textbf{20.2} & \textbf{26.1} & 32.4 \\
    \midrule
    \multirow{9}{*}{ResNet-101} & \textit{two-stage} &  &  &  &  &  &  &  \\
    ~ & Faster R-CNN \cite{ren2015faster} & end-to-end & RFS & 24 & 25.7 & 15.1 & 24.1 & 32.0 \\
    ~ & EQLv2 \cite{tan2021equalization} & end-to-end & RFS & 24 & 26.9 & 18.2 & 25.4 & 32.4 \\
    ~ & Seasaw Loss \cite{wang2021seesaw} & end-to-end & RFS & 24 & 27.8 & 18.7 & 27.0 & 32.8 \\
    ~ & BAGS \cite{li2020overcoming} & decoupled & RFS+CBS & 24+12 & 27.6 & 18.7 & 26.5 & 32.6 \\
    \cmidrule{2-9}
    ~ & \textit{one-stage} &  &  &  &  &  &  &  \\
    ~ & RetinaNet \cite{lin2017focal} & end-to-end & RFS & 24 & 19.6 & 10.1 & 17.3 & 26.2 \\
    ~ & Baseline$^{\dag}$ & end-to-end & RFS & 24 & 27.0 & 14.4 & 25.7 & \textbf{34.0} \\
    ~ & EFL (Ours) & end-to-end & RFS & 24 & \textbf{29.2} & \textbf{23.5} & \textbf{27.4} & 33.8 \\
    \bottomrule
  \end{tabular}}
  \caption{Main results of EFL compared with other methods on LVIS v1 \texttt{val} split.
           Baseline$^{\dag}$ indicates the improved baseline.
           RFS and CBS indicate the repeat factor sampler and the class balanced sampler, respectively.
           All end-to-end methods are trained by a schedule of 2x with the RFS while the decoupled methods have an additional 1x schedule with the CBS during the fine-tuning stage.}
  \label{tab:main_results}
\end{table*}

\noindent
\textbf{Implementation Details.} We adopt the ImageNet \cite{deng2009imagenet} pre-trained ResNet-50 \cite{he2016deep} as the backbone with a Feature Pyramid Network (FPN) \cite{lin2017feature} as the neck.
The network is trained using the SGD algorithm with a momentum of 0.9 and a weight decay of 0.0001.
During the training phase, scale jitter and random horizontal flipping are adopted as the data augmentation.
We train the model with a total batch size of 16 on 16 GPUs (1 image per GPU) and set the initial learning rate as 0.02.
The prior probability of the last layer in the classification branch is initialized to 0.001 as mentioned in \cite{lin2017focal, tan2020equalization}.
During the inference phase, we resize the shorter edge of the input image to 800 pixels and keep the longer edge smaller than 1333 pixels without changing the aspect ratio.
No test time augmentation is used.
As one-stage detectors often predict boxes with low scores, we do not filter out any predicted box before NMS (set the minimum score threshold to 0).
Following \cite{gupta2019lvis}, we select the top 300 confident boxes per image as the final detection results.
Since most experimental results of two-stage approaches are based on the Mask R-CNN \cite{he2017mask} framework, their released detection performance AP$_b$ is affected by the segmentation performance.
We report the detection results of those works by rerunning their code with the Faster R-CNN \cite{ren2015faster} framework for a fair comparison.
All models are trained with the repeat factor sampler (RFS) by a 2x schedule.
For our proposed EFL, we set the balanced factor $\alpha_{\mathrm{t}} = 0.25$ and the base focusing parameter $\gamma_b = 2.0$ which are consistent with focal loss \cite{lin2017focal}.
The hyper-parameter $s$ is set to 8 and more details about the impact of such a scale factor is showcased in \cref{sec:ablation_studies}.

\noindent
\textbf{Stabilized Setting and Improved Baseline.} As presented in \cref{tab:eqlv2_baseline}, there is a large performance gap between the two-stage baseline Faster R-CNN \cite{ren2015faster} and the widely-used one-stage detector RetinanNet \cite{lin2017focal}.
To bridge this gap, we investigate plenty of one-stage frameworks to establish an improved baseline that is more appropriate for the long-tailed task.
ATSS \cite{zhang2020bridging} stands out among those methods with its simplicity and high performance.
From experiments, we discover that the training processes of most one-stage detectors are pretty unstable with result fluctuations and sometimes encountering NaN problems.
Intuitively, the primary culprit is the abnormal gradients in the early training stage caused by severe imbalance problems.
To stabilize the training processes, we adopt the stabilized settings that extend the warm-up iterations from 1000 to 6000 and utilize the gradient clipping with a maximum normalized value of 35.
Meanwhile, in ATSS, we adopt an IoU branch to replace the centerness branch and set the anchor scale from \{8\} to \{6, 8\} with the hyper-parameter $k=18$ to cover more potential candidates.
The combination of stabilized and improved settings is adopted as our improved baseline.
Unless otherwise stated, EFL is trained with the improved baseline.

\subsection{Benchmark Results}

To show the effectiveness of our proposed method, we compare our approach with other works that report state-of-the-art performance.
As demonstrated in \cref{tab:main_results}, with ResNet-50-FPN backbone, our proposed method achieves an overall 27.5\% AP, which improves the proposed improved baseline by 1.8\% AP, and even achieves 5.9 points improvement on rare categories.
The result shows that EFL could handle the extreme positive-negative imbalance problem of rare categories well.
Compared with other end-to-end methods like EQL \cite{tan2020equalization}, EQLv2 \cite{tan2021equalization}, and Seesaw Loss \cite{wang2021seesaw}, our proposed method outperforms them by 2.4\% AP, 2.0\% AP, and 1.1\% AP, respectively.
And compared with decoupled training approaches like cRT \cite{kang2019decoupling} and BAGS \cite{li2020overcoming}, our approach surpasses them with an elegant end-to-end training strategy (by 2.7\% AP and 1.5\% AP).
Besides the high performance, we keep the advantages of one-stage detectors like simplicity, rapidity, and ease of deployment.

With the larger ResNet-101-FPN backbone, our approach still performs well on the improved baseline (+2.2\% AP).
Meanwhile, our approach maintains stable performance improvements compared with all existing methods, whether they are end-to-end or decoupled.
Without bells and whistles, our method achieves 29.2\% AP that establishes a new state-of-the-art.
Notably, the performance of rare categories on the improved baseline does not gain too much performance improvement from the larger backbone while EFL does (from 20.2\% AP to 23.5\% AP).
It indicates that our proposed method has a good generalization ability across different backbones.

\subsection{Ablation Studies}
\label{sec:ablation_studies}

\noindent
\textbf{Influence of components in EFL.} There are two components in EFL, which are the focusing factor and the weighting factor.
To demonstrate the effect of each component, we train the model with our proposed improved baseline by a 2x schedule and repeat factor sampler.
As shown in \cref{tab:ablation_study_components}, both the weighting factor and the focusing factor play significant roles in EFL.
For the focusing factor, it achieves an improvement from 25.7\% AP to 26.2\% AP.
Meanwhile, it brings a significant gain on rare categories with 3.4\% AP improvement, indicating its effectiveness in alleviating the severe positive-negative imbalance problems.
And for the weighting factor, we investigate its influence by setting the focusing factor always be $\gamma_b$ among all categories in EFL.
Thus the function of the weighting factor could also be regarded as a reweighting approach combined with focal loss.
As expected, the weighting factor outperforms the improved baseline by 0.4\% AP.
By the synergy of these two components, EFL dramatically improves the performance of improved baseline from 25.7\% AP to 27.5\% AP.

\begin{table}
  \centering
  \setlength{\tabcolsep}{3.5mm}{
  \begin{tabular}{cc|cccc}
    \toprule
    WF & FF & AP & $AP_r$ & $AP_c$ & $AP_f$ \\
    \midrule
     &  & 25.7 & 14.3 & 23.8 & \textbf{32.7} \\
    \checkmark &  & 26.1 & 15.6 & 24.5 & 32.6 \\
     & \checkmark & 26.2 & 17.7 & 24.7 & 31.5 \\
    \checkmark & \checkmark & \textbf{27.5} & \textbf{20.2} & \textbf{26.1} & 32.4 \\
    \bottomrule
  \end{tabular}}
  \caption{Ablation study of each component in Equalized Focal Loss.
  WF, FF indicate the weighting factor and the focusing factor, respectively.}
  \label{tab:ablation_study_components}
\end{table}

\begin{table}
  \centering
  \setlength{\tabcolsep}{5mm}{
  \begin{tabular}{c|cccc}
    \toprule
    $s$ & AP & AP$_r$ & AP$_c$ & AP$_f$ \\
    \midrule
    0 & 25.7 & 14.3 & 23.8 & \textbf{32.7} \\
    1 & 26.3 & 16.3 & 24.6 & 32.6 \\
    2 & 26.6 & 17.6 & 24.6 &  \textbf{32.7} \\
    4 & 27.3 & 19.9 & 25.5 & 32.6 \\
    8 & \textbf{27.5} & \textbf{20.2} & \textbf{26.1} & 32.4 \\
    12 & 26.5 & 19.9 & 24.6 & 31.6 \\
    \bottomrule
  \end{tabular}}
  \caption{Ablation study of the hyper-parameter $s$.
           $s=8$ is adopted as the default setting in other experiments.}
  \label{tab:s_effect}
\end{table}

\noindent
\textbf{Influence of the Hyper-parameter.} Too many hyper-parameters will affect the generalization ability of a method.
In this paper, our proposed EFL only has one hyper-parameter $s$ that is also one of the advantages of our work.
We study the influence of $s$ with different values and find that $s=8$ achieves the best performance.
As presented in \cref{tab:s_effect}, almost all $s \geq 0$ could increases the performance of the improved baseline.
What's more, $s$ keeps relatively high performance in a wide range.
It indicates that our EFL is hyper-parameter insensitive.

\begin{table}
  \centering
  \setlength{\tabcolsep}{3mm}{
  \begin{tabular}{ccc|cccc}
    \toprule
    STS & IB & IA & AP & AP$_r$ & AP$_c$ & AP$_f$ \\
    \midrule
     &  &  & 25.8 & 18.1 & 24.5 & 30.6 \\
    \checkmark &  &  & 26.1 & 18.6 & 24.8 & 30.8 \\
    \checkmark & \checkmark &  & 26.8 & 18.7 & 25.4 & 31.9 \\
    \checkmark &  & \checkmark & 26.5 & 20.0 & 24.8 & 31.1 \\
    \checkmark & \checkmark & \checkmark & \textbf{27.5} & \textbf{20.2} & \textbf{26.1} & \textbf{32.4} \\
    \bottomrule
  \end{tabular}}
  \caption{Ablation study of each component in our proposed improved baseline.
  STS, IB, IA indicate the stabilized settings, IoU branch, and increased anchor scale, respectively.}
  \label{tab:improved_baseline}
\end{table}

\noindent
\textbf{Influence of components in the improved baseline.} We investigate the influence of three components in our proposed improved baseline: the stabilized settings, the IoU branch, and the increased anchor scale.
The models are all trained with our proposed EFL.
As presented in \cref{tab:improved_baseline}, the stabilized settings stabilize the training process and bring some performance improvement (+0.3\% AP). Combined with the stabilized settings, the IoU branch and the increased anchor scale improve the original EFL by 1.0\% AP and 0.7\% AP, respectively.
Notably, compared with the ATSS in \cref{tab:eqlv2_baseline}, even without the stabilized and improved settings, our EFL still brings a significant performance improvement (from 24.7\% AP to 25.8\% AP, +1.1\% AP).
And combined with those settings, the performance improvement is further increased.
Our approaches outperform the improved baseline by 1.8\% AP as demonstrated in \cref{tab:main_results}.
The results indicate that the improved baseline is beneficial to the improvement of EFL but not necessary.
Meanwhile, the stabilized settings are detector-agnostic that could be applied well with other one-stage detectors.
Actually, all RetinaNet and FCOS$^*$ experiments are trained with the stabilized settings to avoid the NaN problem.

\subsection{Model Analysis}
\label{sec:model_analysis}

\begin{table}
  \centering
  \setlength{\tabcolsep}{1.5mm}{
  \begin{tabular}{l|c|cccc}
    \toprule
    method & EFL & AP & $AP_r$ & $AP_c$ & $AP_f$ \\
    \midrule
    \multirow{2}{*}{RetinaNet \cite{lin2017focal}} &  & 18.5 & 9.6 & 16.1 & \textbf{25.0} \\
    ~ & \checkmark & \textbf{20.5 (+2.0)} & \textbf{15.2} & \textbf{18.6} & 24.8 \\
    \midrule
    \multirow{2}{*}{FCOS$^*$ \cite{tian2019fcos}} &  & 22.6 & 12.7 & 20.9 & \textbf{28.9} \\
    ~ & \checkmark & \textbf{23.4 (+0.8)} & \textbf{14.9} & \textbf{21.9} & 28.7 \\
    \midrule
    \multirow{2}{*}{PAA \cite{kim2020probabilistic}} &  & 23.7 & 14.2 & 21.6 & 30.2 \\
    ~ & \checkmark & \textbf{25.6 +(1.9)} & \textbf{19.8} & \textbf{23.8} & \textbf{30.2} \\
    \midrule
    \multirow{2}{*}{Baseline$^{\dag}$} &  & 25.7 & 14.3 & 23.8 & \textbf{32.7} \\
    ~ & \checkmark & \textbf{27.5 +(1.8)} & \textbf{20.2} & \textbf{26.1} & 32.4 \\
    \bottomrule
  \end{tabular}}
  \caption{Results of EFL combined with other one-stage object detectors.
           FCOS$^{*}$ indicates that the reported FCOS result is trained with a center-sampling strategy \cite{zhang2020bridging}.
           Baseline$^{\dag}$ indicates our proposed improved baseline.}
  \label{tab:one_stage_with_efl}
\end{table}

\noindent
\textbf{Combined with other one-stage detectors.} To demonstrate the generalization ability of EFL across different one-stage detectors, we combine it with RetinaNet, FCOS$^*$, PAA, and our improved baseline, separately.
As presented in \cref{tab:one_stage_with_efl}, EFL performs well with all those one-stage detectors.
Compared with the combination of EQLv2 and focal loss in \cref{tab:eqlv2_baseline}, our proposed EFL maintains a stable big performance improvement (about +2\% AP) to the original detector.
What's more, EFL dramatically improves the performance of those detectors on rare categories, which shows our strength in settling the long-tailed distribution problem.

\begin{figure}
  \centering
  \includegraphics[width=1.0\linewidth]{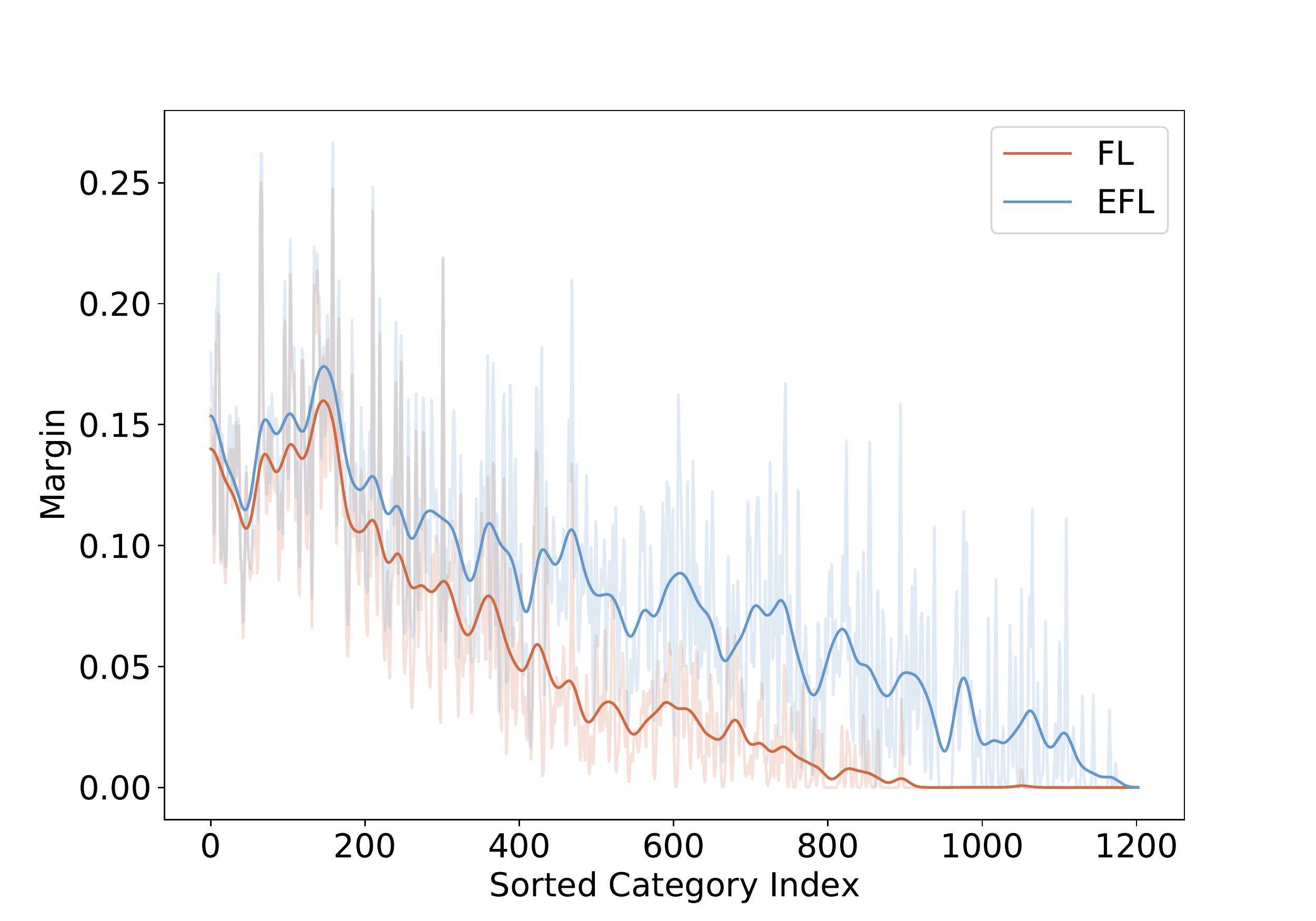}
  \caption{
  Comparisons of the margins among categories of the focal loss and EFL on LVIS v1 \texttt{val} split.
  The margin is calculated by subtracting the mean predicted score of negative samples from the mean predicted score of positive samples for each category.
  A larger margin indicates that the classifier has a more distinct decision boundary.}
  \label{fig:score_margin}
\end{figure}

\noindent
\textbf{A more distinct decision boundary.} We investigate whether EFL has more distinct decision boundaries than the focal loss on rare categories classifiers.
Since the decision boundaries cannot be displayed intuitively, the empirical margins between positive samples and negative samples among categories are adopted to reflect those boundaries.
The margin is calculated by subtracting the mean predicted score of negative samples from the mean predicted score of positive samples for each category.
As EFL greatly focuses learning on the positive-negative imbalance problem of rare categories, the margins between positive samples and negative samples of those categories in EFL should be prominent.
As presented in \cref{fig:score_margin}, the baseline focal loss keeps small margins between positive samples and negative samples on rare categories.
In contrast, our proposed EFL increases the margins of all categories especially for rare categories, resulting in more distinct decision boundaries.

\begin{table}
  \centering
  \setlength{\tabcolsep}{1mm}{
  \begin{tabular}{l|c|ccccc}
    \toprule
    method & AP & AP$1$ & AP$2$ & AP$3$ & AP$4$ & AP$5$ \\
    \midrule
    \textit{R-50 w/ FPN}  &  &  &  &  &  &  \\
    Faster R-CNN \cite{ren2015faster} & 43.1 & 26.3 & 42.5 & 45.2 & 48.2 & \textbf{52.6} \\
    RetinaNet \cite{lin2017focal} & 32.1 & 21.0 & 34.0 & 35.4 & 35.6 & 34.2 \\
    Baseline$^{\dag}$ \cite{zhang2020bridging} & 43.3 & 19.4 & 44.3 & 49.5 & \textbf{50.6} & 52.2 \\
    \textbf{EFL (Ours)} & \textbf{51.5} & \textbf{52.8} & \textbf{52.9} & \textbf{50.8} & 50.2 & 50.9 \\
    \midrule
    \textit{R-101 w/ FPN}  &  &  &  &  &  &  \\
    Faster R-CNN \cite{ren2015faster} & 46.0 & 29.2 & 45.5 & 49.3 & 50.9 & \textbf{54.7} \\
    RetinaNet \cite{lin2017focal} & 35.8 & 26.4 & 38.9 & 38.3 & 38.1 & 36.8 \\
    Baseline$^{\dag}$ \cite{zhang2020bridging} & 44.7 & 19.6 & 46.6 & 51.1 & \textbf{52.3} & 53.4 \\
    \textbf{EFL (Ours)} & \textbf{52.6} & \textbf{53.4} & \textbf{53.8} & \textbf{51.4} & 51.8 & 52.3 \\
    \bottomrule
  \end{tabular}}
  \caption{Results on OpenImages Challenge 2019 \texttt{val} split.
  Baseline$^{\dag}$ indicates our proposed improved baseline.}
  \label{tab:oid_results}
\end{table}

\subsection{Results on Open Images Detection}
\label{sec:oid_results}

To verify the generalization ability to other datasets, we conduct a series of experiments on the challenging OpenImage dataset \cite{kuznetsova2020open}.
We adopt the Challenge 2019 split as the benchmark.
Challenge 2019 is a subset of OpenImages V5, and it contains 500 categories that also suffer from the long-tailed distribution problem.
The 1.7M images in the \texttt{train} split are used for training while the 41k \texttt{val} split is used for validation.
It is worth noting that all experimental settings are consistent with them searched in the LVIS v1 benchmark without any tuning.
To better understand the improvement on rare categories, following the settings of \cite{tan2020equalization, tan2021equalization}, the categories in Challenge 2019 are divided into five groups (100 categories per group) according to the instance numbers.
We adopt the widely-used mAP@IoU=0.5 metric for evaluation and indicate the mAP of the above five groups as AP1 to AP5.
Where AP1 is the AP of the first group with the rarest categories and AP5 is the AP of the last group with the most frequent categories.
All models are trained by a schedule of 120k/160k/180k with the random sampler.
As presented in \cref{tab:oid_results}, our proposed improved baseline greatly improves the performance of one-stage detectors.
It brings about 11\% AP improvement compared with the widely used RetinaNet.
Combined with the improved baseline, our proposed EFL achieves an overall AP of 51.5\% with the ResNet-50 backbone, which outperforms the two-stage baseline Faster R-CNN and the improved baseline by 8.4\% AP and 8.2\% AP, respectively.
What's more, EFL significantly improves the performance of the rare categories with an improvement of 33.4\% AP on AP1 split. 
With the larger ResNet-101 backbone, our proposed method still performs well and brings significant AP gains.
Meanwhile, it maintains an excellent performance on rare categories.
All experimental results demonstrate the strength and generalization ability of our method.

\section{Conclusion}

In this work, we study how to build a high-performance one-stage object detector in the long-tailed case. We identify the positive-negative imbalance degree inconsistency among categories as the primary difficulty. A novel Equalized Focal Loss (EFL) is proposed to protect the learning of one-stage detectors from extreme imbalance problems. Our proposed EFL is the first solution to the one-stage long-tailed object detection. Combined with some improved techniques and stabilized settings, a strong one-stage detector with EFL beats all existing state-of-the-art methods on the challenging LVIS v1 benchmark.

\appendix

\begin{table}
  \centering
  \setlength{\tabcolsep}{0.8mm}{
  \begin{tabular}{l|c|c|cccc}
    \toprule
    model & loss & YOLOX$^{*}$ & AP & AP$_r$ & AP$_c$ & AP$_f$ \\
    \midrule
    \multirow{5}{*}{small} & Sigmoid &  & 15.2 & 2.9 & 11.6 & 24.7 \\
    \cmidrule{2-7}
    ~ & FL & \checkmark & 18.5 & 3.6 & 15.7 & \textbf{28.2} \\
    ~ & \textbf{EFL(Ours)} & \checkmark & \textbf{23.3} & \textbf{18.1} & \textbf{21.2} & 28.0 \\
    \cmidrule{2-7}
    ~ & QFL & \checkmark & 22.5 & 11.0 & 20.6 & \textbf{29.7} \\
    ~ & \textbf{EQFL(Ours)} & \checkmark & \textbf{24.2} & \textbf{16.3} & \textbf{22.7} & 29.4 \\
    \midrule
    \multirow{5}{*}{medium} & Sigmoid &  & 20.9 & 5.3 & 17.6 & 31.5 \\
    \cmidrule{2-7}
    ~ & FL & \checkmark & 25.0 & 7.1 & 23.5 & 34.4 \\   
    ~ & \textbf{EFL(Ours)} & \checkmark & \textbf{30.0} & \textbf{23.8} & \textbf{28.2} & \textbf{34.7} \\
    \cmidrule{2-7}
    ~ & QFL & \checkmark & 28.9 & 16.8 & 27.2 & 36.1 \\
    ~ & \textbf{EQFL(Ours)} & \checkmark & \textbf{31.0} & \textbf{24.0} & \textbf{29.1} & \textbf{36.2} \\
    \bottomrule
  \end{tabular}}
  \caption{Results of the YOLOX \cite{ge2021yolox} detectors on the LVIS v1 \cite{gupta2019lvis} dataset.
  All experiments are trained from scratch by 300 epochs with the repeat factor sampler (RFS).
  The YOLOX$^*$ indicates the enhanced YOLOX detector that is trained with our proposed improved settings.
  FL and QFL indicate the focal loss and the quality focal loss, respectively.
  EFL and EQFL are the methods proposed in this paper that indicate the equalized version of FL and QFL.
  The scale factor $s$ in EFL and EQFL are all set to 4.}
  \label{tab:yolox_results}
\end{table}

\section{Combined with YOLOX}

YOLOX \cite{ge2021yolox} is a recently proposed one-stage detector based on the YOLO series.
Its remarkable performance and extremely fast inference speed have won the favor of researchers and developers.
In this paper, we investigate whether the advanced YOLOX detector works well under the long-tailed data distribution.
Then we introduce our proposed method into the YOLOX detector to help it achieve excellent performance.
The experiments are conducted on the small and medium models of YOLOX (YOLOX-S and YOLOX-M).
The challenging LVIS v1 dataset is adopted as the benchmark.
All networks are trained from scratch by 300 epochs with the repeat factor sampler (RFS).
Unless otherwise stated, our experimental settings are aligned with the original settings in \cite{ge2021yolox}\footnote[2]{We highly recommend readers to refer to \url{https://github.com/Megvii-BaseDetection/YOLOX} for more details.}.

As presented in \cref{tab:yolox_results}, both YOLOX-S and YOLOX-M perform poorly in the long-tailed scenario.
We argue that the poor performance mainly comes from two aspects.
On the one hand, the supervisor (especially the OTA \cite{ge2021ota} label assignment strategy) in the YOLOX detector is influenced by the long-tailed data distribution which results in low-quality supervision during the training phase.
On the other hand, the classification loss in the YOLOX detectors is the sigmoid loss which is incapable to handle the severe positive-negative imbalance degree inconsistency problem (see \cref{sec:efl_definition} for more details).
Based on these analyses, we make some modifications to the YOLOX detector and apply our proposed method to it.
With the following settings, the medium model of YOLOX even achieves an overall AP of 31.0\% which indicates the effectiveness of our method:

\noindent
\textbf{Enhancements on the YOLOX.} Firstly, we replace the supervisor and the predictor of the YOLOX with the settings in our improved baseline (including the ATSS label assignment strategy, IoU branch, and increased anchor scale).
The IoU loss combined with the L1 loss is adopted as the localization loss (YOLOX has the same behavior during the last 15 training epochs).
We denote the YOLOX detector combined with our improved settings as the YOLOX$^*$ series.
As shown in \cref{tab:yolox_results}, with these enhancements, the YOLOX$^*$ outperforms the YOLOX by a large margin (from 20.9\% AP to 25.0\% AP on the medium model) which indicates that the supervision in the YOLOX$^*$ is more reliable than the YOLOX.

\noindent
\textbf{Adapt EFL to the YOLOX$^*$.} Although the YOLOX$^*$ performs better than the YOLOX, its training process is still highly biased towards the frequent categories (AP$_r$ is only 7.1\% on the medium model).
Thus we adapt our proposed EFL to the YOLOX$^*$ to address the long-tailed imbalance issues.
It is worth noting that we empiracally set the weight decays of bias parameters in the last layer of the classification head to 0.0001 (original setting in the YOLOX is 0), because we discover from experiments that this setting is of vital importance on the performance of EFL.
Without this slight modification, the gradient collection mechanism in EFL will malfunction.
The hyper-parameter $s$ in our EFL is set to 4 when applied to the YOLOX$^*$.
As presented in \cref{tab:yolox_results}, combined with the YOLOX$^*$ series, EFL achieves excellent performance in the long-tailed situation.
On the medium model, it reaches an overall AP of 30.0\% that outperforms the YOLOX-M$^*$ detector by 5.0\% AP.
What's more, it greatly improves the performance of the rare categories with +16.7\% AP.
The results demonstrate that our proposed EFL is a very practical approach that could greatly alleviate the long-tailed imbalance problem for almost all one-stage detectors.

\noindent
\textbf{Equalized Quality Focal Loss.} 
Meanwhile, we also investigate the performance of the quality focal loss (QFL) \cite{li2020generalized} combined with the YOLOX$^*$ series.
It could be concluded from experiments that QFL achieves more competitive results compared with the focal loss.
We wonder whether the performance of QFL could be further improved by drawing ideas from the EFL.
Then the class-relevant modulating factor is designed for the QFL and we denote the novel loss as the equalized quality focal loss (EQFL).
The EQFL of the $j$-th category is formulated as:
\begin{equation}
  \mathrm{EQFL}\left(p\right)=-m^j_f\left(y^{\prime}\log\left(p\right) + \left(1-y^{\prime}\right)\log\left(1-p\right)\right)
  \label{eq:eqfl_loss}
\end{equation}
where $m^j_f=w^j_f\left(\left|y^{\prime}-p\right|\right)^{f^j_f}$ is the specific form of the modulating factor in EQFL.
The weighting factor and the focusing factor (\eg the $w^j_f$ and the $f^j_f$) are the same as them in EFL.
It should be noticed that $y{\prime} \in [0, 1]$ here is the IoU score for a positive sample and 0 for a negative sample as declared in \cite{li2020generalized}.
Our proposed EQFL achieves 31.0\% AP on the medium model.
We hope that the impressive performance and powerful generalization ability of our proposed method could inspire the community to raise more attention on one-stage detectors in the long-tailed case.

\begin{figure}
  \centering
  \includegraphics[width=1.0\linewidth]{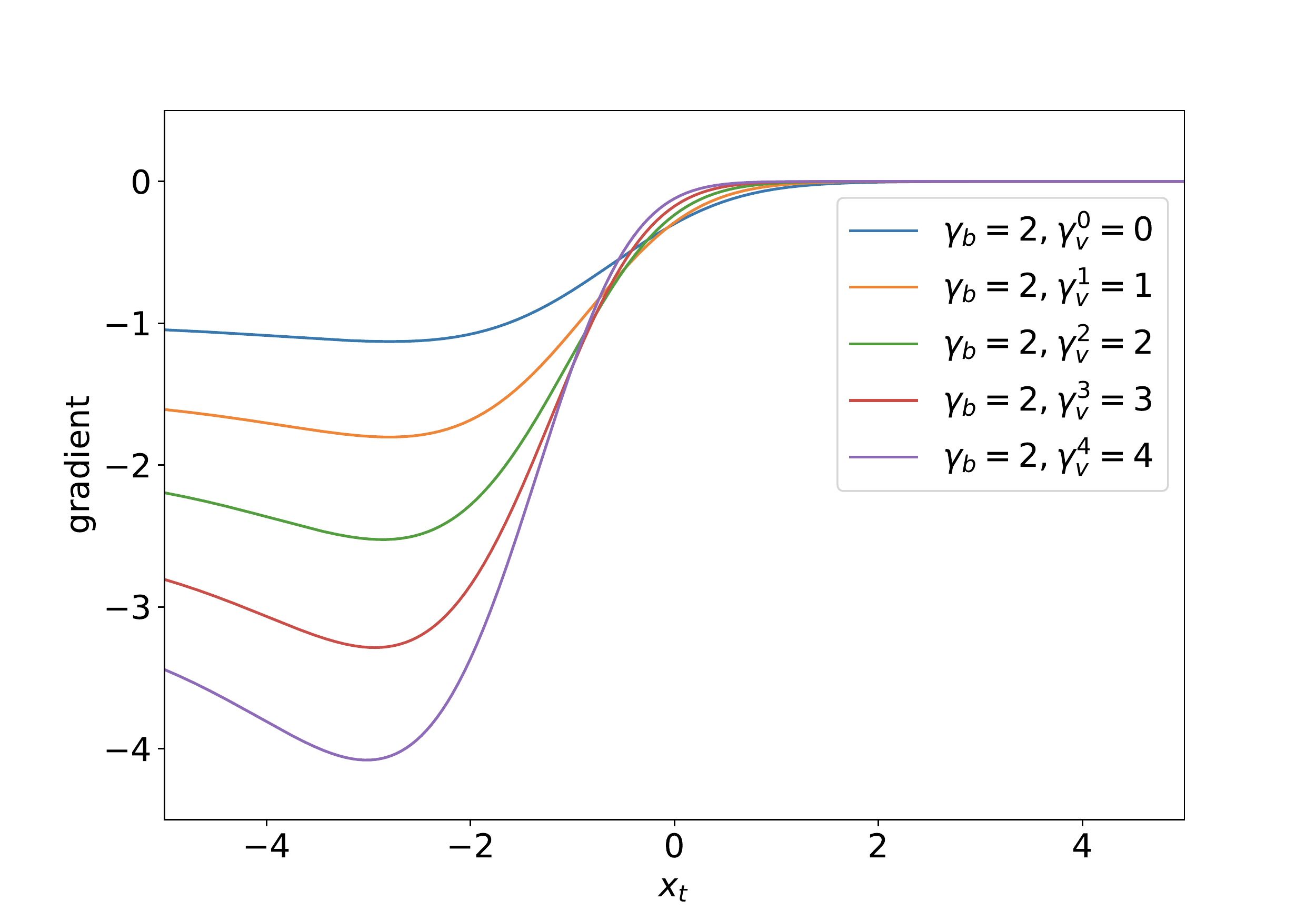}
  \caption{Derivative of our proposed EFL.
  $x_{\mathrm{t}} = (2y - 1) x$, where $x$ is the output predicted logit and $y \in \{0, 1\}$ specific the ground-truth label of the binary classification.
  In this figure, different colors indicate different categories.
  We set $\gamma_b = 2$ and ignore the impact of $\alpha_{\mathrm{t}}$.}
  \label{fig:derivative}
\end{figure}

\section{Derivative}

The derivative is a crucial part of the parameter $g^j$ in EFL.
It could be used to calculate the accumulated gradient of positive samples and negative samples.
For reference, the derivative for EFL of the $j$-th category is:
\begin{equation}
  \frac{\mathrm{d}\mathrm{EFL}}{\mathrm{d}x}=\frac{\gamma^j \left(2y - 1\right)}{\gamma_b}\left(1-p_{\mathrm{t}}\right)^{\gamma^j}\left(\gamma^j p_{\mathrm{t}} \log \left(p_{\mathrm{t}}\right) + p_{\mathrm{t}} - 1\right)
  \label{eq:efl_derivatives}
\end{equation}
where $y \in \{0, 1\}$ specific the ground-truth label of the binary classification.
Plots for the derivatives of different categories are shown in \cref{fig:derivative}.
For all categories, EFL has small derivatives for easy samples ($x_{\mathrm{t}} > 0$).
As $\gamma_v^j$ increases (\eg the category becomes rare), EFL gradually improves the gradient contribution of hard samples, resulting in more concentration on learning them. 

In addition to manually calculating the gradients, an alternate approach is to register a backward hook on the classification loss function to get the gradients of positive and negative samples.
Meanwhile, when adapting EQLv2 \cite{tan2021equalization} to one-stage detectors, we calculate the gradients of the EQLv2\&Focal.
EQLv2\&Focal is the combination of EQLv2 and focal loss.
It directly applies the gradient-guided reweighting mechanism to the focal loss.
Here we also show the derivative for EQLv2\&Focal:
\begin{equation}
  \frac{\mathrm{d}f\left(x\right)}{\mathrm{d}x}=w_{\mathrm{t}}\left(2y - 1\right)\left(1-p_{\mathrm{t}}\right)^{\gamma}\left(\gamma p_{\mathrm{t}} \log \left(p_{\mathrm{t}}\right) + p_{\mathrm{t}} - 1\right)
  \label{eq:eqlv2_derivatives}
\end{equation}
where $w_{\mathrm{t}}$ indicates the gradient-guided weight to the focal loss.
It is worth noting that the backward hook is the same applies in this situation.

\begin{table}
  \centering
  \setlength{\tabcolsep}{1.5mm}{
  \begin{tabular}{l|ccc|ccc}
    \toprule
    loss & AP & AP$_{50}$ & AP$_{75}$ & AP$_s$ & AP$_m$ & AP$_l$ \\
    \midrule
    FL & 42.3 & 61.0 & 45.7 & 26.7 & 46.2 & 53.0  \\
    EFL(s=2) & 42.4 & 61.0 & 46.1 & 26.5 & 46.2 & 53.2 \\
    EFL(s=4) & 42.3 & 61.2 & 45.6 & 26.6 & 46.1 & 52.9 \\
    EFL(s=8) & 42.2 & 60.8 & 45.6 & 26.4 & 46.1 & 52.7 \\
    \bottomrule
  \end{tabular}}
  \caption{Results in the COCO dataset.
  All results are came from the improved baseline with the ResNet-50 backbone.
  The models are trained by a 2x schedule with the random sampler.}
  \label{tab:coco_results}
\end{table}

\section{Performance on COCO Dataset}
As we claim in this paper, EFL is equivalent to the focal loss in the balanced data scenario.
To verify this analysis, we conduct experiments on MS COCO dataset \cite{lin2014microsoft}.
COCO is a widely used object detection dataset that includes 80 categories with a balanced data distribution.
The ImageNet pre-trained ResNet-50 is adopted as the backbone, and the networks are trained with our proposed improved baseline.
We train the focal loss and our proposed EFL by a 2x schedule with the random sampler.
All other settings are consistent with those in LVIS.

As presented in \cref{tab:coco_results}, the scaling factor $s$ has little effect in the COCO dataset, and all results with EFL achieve comparable performance with the focal loss.
This indicates that our proposed EFL could maintain good performance under the balanced data distribution.
EFL does not rely on pre-computing the distribution of training data and could operate well with any data sampler.
This distribution-agnostic property enables EFL to work well with real-world applications in different data distributions.

{\small
\bibliographystyle{ieee_fullname}
\bibliography{reference}
}

\end{document}